\documentclass[9pt,twocolumn,twoside]{pnas-new}
\usepackage{hyperref,lineno}
\usepackage[utf8]{inputenc}
\usepackage[T1]{fontenc}

\templatetype{pnasresearcharticle} 

\title{Restoration of Fragmentary Babylonian Texts Using Recurrent Neural Networks}
\usepackage{todonotes}
\author[a,1,2]{Ethan Fetaya}
\author[b]{Yonatan Lifshitz} 
\author[c]{Elad Aaron}
\author[c,1,2]{Shai Gordin}

\affil[a]{Faculty of Engineering, Bar-Ilan University}
\affil[b]{Alpha Program, Davidson Institute of Science Education, Weizmann Institute of Science}
\affil[c]{Israel Heritage Department and the Institute of Archaeology, Ariel University}

\leadauthor{Fetaya} 

\authorcontributions{E.F. and Y.L. collected data, performed tokenization, designed the algorithms and trained them. E.A. prepared the text YOS 7 11 for publication, went over data points and training questions. Sh.G. guided the research and model training, went over published texts and prepared YOS 7 51. Sh.G. and E.F. wrote the paper together and headed the research team.}
\authordeclaration{The authors declare no conflict of interest.}
\equalauthors{\textsuperscript{1}E.F. and Sh.G. contributed equally to this work.}
\correspondingauthor{\textsuperscript{2}To whom correspondence should be addressed. E-mail: Ethan.Fetaya@biu.ac.il, shaigo@ariel.ac.il}

\keywords{Babylonian heritage $|$ cuneiform script $|$ late Babylonian dialect $|$ Achaemenid empire $|$ neural networks} 

\begin{abstract}
The main source of information regarding ancient Mesopotamian history and culture are clay cuneiform tablets. Despite being an invaluable resource, many tablets are fragmented leading to missing information. Currently these missing parts are manually completed by experts. In this work we investigate the possibility of assisting scholars and even automatically completing the breaks in ancient Akkadian texts from Achaemenid period Babylonia by modelling the language using recurrent neural networks. 
\end{abstract}

\dates{This manuscript was compiled on \today}
\doi{\url{www.pnas.org/cgi/doi/10.1073/pnas.XXXXXXXXXX}}

\begin{document}
\nolinenumbers
\maketitle
\begin{figure}[ht] 
\centering  
\includegraphics[width=0.5\textwidth]{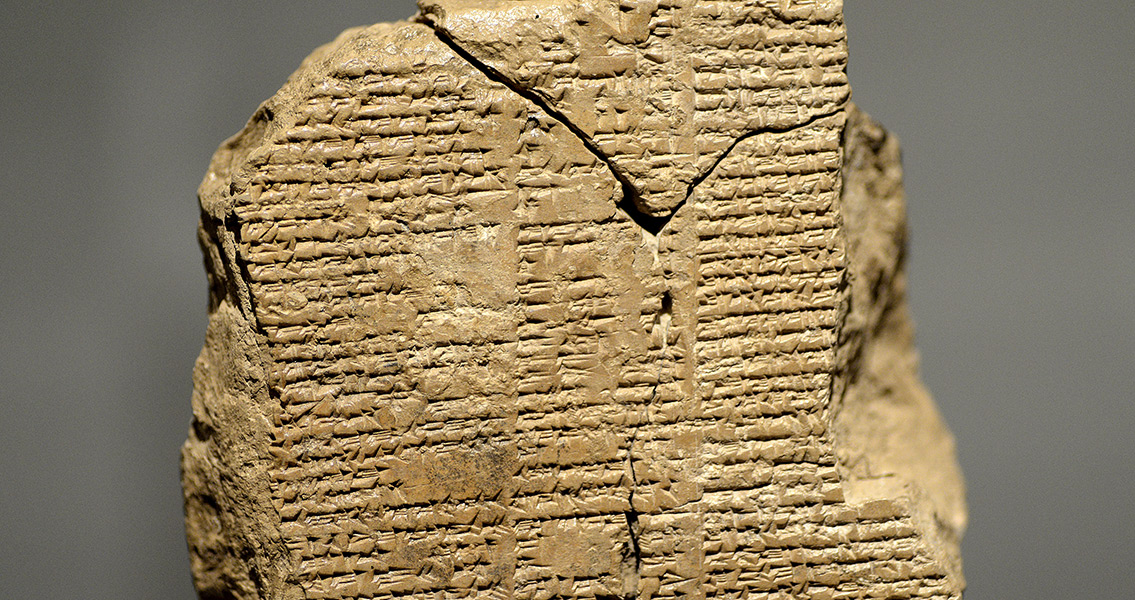}
\caption{Example of a typical fragmentary clay tablet in photo. The text is the fifth tablet in the famous Epic of Gilgameš, Suleimaniyah Museum T.1447. Photo © Osama S. M. Amin.}
\label{fig:tablet}
\end{figure}

\thispagestyle{firststyle}
\ifthenelse{\boolean{shortarticle}}{\ifthenelse{\boolean{singlecolumn}}{\abscontentformatted}{\abscontent}}{}
\dropcap{C}uneiform is one of the two earliest forms of writing known in mankind's history (the other being Egyptian Hieroglyphs). It was used to write one of the main languages of the ancient world, Akkadian, from the Semitic language family. More than 2,500 years of human activity has been recorded in several dialects of this language across most of the ancient Near East \cite{HuehnergardWoods04, Streck11}. In all, more than 10 million words are attested across some 600,000 inscribed clay tablets and hundreds of monumental inscriptions, most of them on stone, that are kept in various collections around the world \cite{Streck10}. Most importantly, Akkadian is our main, and sometimes the only, cultural source regarding some of the most prominent civilizations of the ancient world. These include the Akkadian Empire of Sargon in the third millennium BCE, the Empires of the Late Bronze Age for which it served as \textit{lingua franca}, and the Neo-Assyrian, Neo-Babylonian and Persian Empires.
Clay tablets, though a rather durable medium, are frequently found in fragmentary state or become brittle and deteriorate when not properly handled once exposed to the elements, after excavation or in later museum storage \cite{Guetschow12}. Cracks, broken off or eroded clay and stone or sometimes completely missing pieces, render difficulties in fully deciphering the information recorded in the inscription (see Fig. \ref{fig:tablet})

Thus, large pieces of our ancient cultural heritage become lost. Current practice is to estimate these missing parts manually. This process requires extensive knowledge in the specific genre and corpus, and is currently done by a handful of experts sifting through a large number of parallel passages.
Such manual process is time consuming, the completions are subjective and there is no way to quantify the uncertainty in each completion. One possible way to ameliorate these issues is to have an automatic process that can aid human experts in this task, or even completely replace them. In this work we investigate this approach, of automatically completing broken late Babylonian archival texts, using modern machine learning methods, specifically recurrent neural networks \cite{deeplearningbook}.
Due to the limited number of digitized texts it is not obvious that such data-driven methods should work well, but we hypothesized that for genres with highly structured syntax--like the legal, economic and administrative late Babylonian texts--these models should work well as we will show in this work. 
\newline
\subsection*{The Neo- and Late Babylonian Corpora}
One challenge is the limited data available in digital form. For similar unsupervised language modeling tasks in English, for example, one can collect practically endless amounts of texts online where the main limitation is the computational challenge in storing and processing large quantities of data \cite{radford2019language}. For cuneiform texts this is not the case, and one has to use limited manually transliterated texts or automatic optical character recognition (OCR) which is still far from perfect \cite{Maraetal10}. In general, the chronological span of Akkadian cuneiform is large and the selection of available genres for study is heterogeneous. Many periods have a limited amount of digital text available and can supply only low amounts of data to train the learning algorithm. The largest number of digital transliteration is available on the \textit{Open Richly Annotated Cuneiform Corpus} (\href {http://oracc.museum.upenn.edu/}{ORACC website}). Out of close to 10,000 cuneiform texts in total, more than 7,000 have lemmatized transliterations with linguistically tagged lemmas in \href {http://oracc.museum.upenn.edu/doc/help/editinginatf/cdliatf/index.html}{ATF encoding technology}; the majority belong to texts from first millennium BCE ancient Near Eastern Empires (\href {http://oracc.museum.upenn.edu/oimea/}{OIMEA website}). Our choice of texts, however, was governed by a corpus based approach, in order to have more control on the diversity of text genres and phrasing. Three corpora stand out as being extraordinarily rich in available digital transliterations: the Old Babylonian (c. 1900-1600 BCE, see \href {http://www.archibab.fr/}{ARCHIBAB website}), the Neo-Assyrian (c. 1000-600 BCE, see \href {http://oracc.museum.upenn.edu/saao/}{SAAo website}) and the Neo- and Late Babylonian (c. 650 BCE-100 CE).

\begin{figure*}[ht]
 \includegraphics[width=17.8cm]{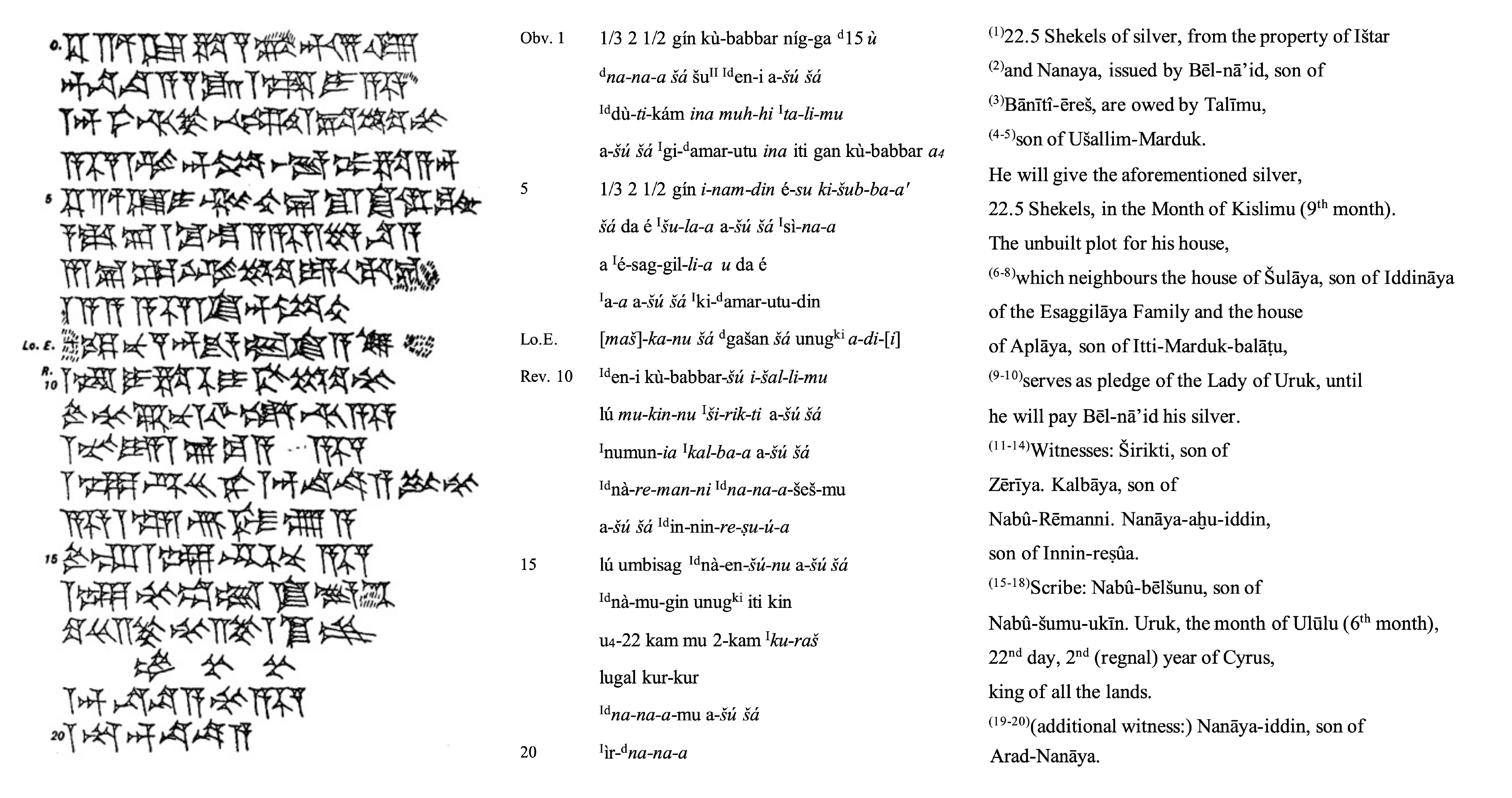}
 \caption{Left to right: The original cuneiform line art, the transliteration and the translation of the Achaemenid period Babylonian text YOS 7 11.}\label{Fig:example}
\end{figure*}

In this work we gather c. 3,000 late Babylonian transliterated texts from Achaemenid period Babylonia (539-331 BCE) in HTML format  from the \href {http://www.achemenet.com/}{Achemenet website}.\footnote{Initiated by Pierre Briant from the Collège de France in the year 2000, it is entirely dedicated to the history, material culture, texts and art of the Achaemenid Empire. The Babylonian text section is administered by the HAROC team of Francis Joannès (UMR ArScAn 7041, CNRS, Nanterre).} This corpus is written in what is commonly termed the Late Babylonian dialect of Akkadian, attested between the rise of the Neo-Babylonian empire in 627 BCE to the end of the use in the cuneiform script around the first century CE \cite{Streck11a}. Though hereafter we will use Neo-Babylonian, which is a better term for the Akkadian dialect of Babylonia for the whole first millennium BCE.\footnote{As shown already by Streck and recently Hackl, an actual sharp distinction between Neo- and Late Babylonian dialects does not physically exist \cite{Hackl18}, see also below.} The largest number of known texts from this period are archival documents belonging to economic, juridical and administrative genres \cite{Jursa04,Jursa05,Jursa10}. The main reason we expect our models to work well on these texts, despite the small amount of data, is that these tablets are official bureaucratic documents, e.g. legal proceedings, receipts, promissory notes, contracts and so on. They are highly structured, mostly short and prefer parataxis over hypotaxis , see Fig. \ref{Fig:example} for an example. These texts contain a lot of patterns, relatively easy for learning algorithms to model yet tedious for humans, making them ideal for our purpose.
  \section{Algorithmic Background}
In this section we will give a very brief introduction to modelling language using recurrent neural networks (RNN), for a more details account see \cite{deeplearningbook}. We can view language as a series of discrete tokens $x_1,...,x_T$ and our goal is to fit a probabilistic model for such sequences, i.e. we wish to find a parametric model that learns the distribution $p(x_1,...,x_T)$ from samples. The first step is to use an autoregressive model, more specifically we use the factorization $p(x_1,...,x_T)=\prod_{t=1}^T p(x_t|x_1,...,x_{t-1})$. What this means is that we can reduce the problem of modeling a full sentence to predicting the next word given the text seen so far. \newline

In recurrent neural networks this autoregressive model $p(x_t|x_1,...,x_{t-1})$ is fitted using a hidden memory. Given the previous hidden memory $h_{t-2}$  the network first updates the memory based on the new input $x_{t-1}$ and then uses the updated memory to predict the next token and then passes the updated memory to the next step. More formally
\begin{align}
    h_{t-1} &= tanh(W_{hh}h_{t-2}+W_{ih}x_{t-1}+b_h) \\
    prob_t  &= softmax(W_{ho}h_{t-1}+b_o)
\end{align}
Where $x_{t-1}$ is a one-hot representation of the input token, $W$'s are linear mappings and $prob_t$ is the vector of probabilities for each possible next token. Given this parametric model, it is trained by maximizing the training log-likelihood to produce the output model. While simple and effective, due to vanishing gradients simple recurrent neural networks have difficulties in modeling long time dependencies, i.e. when the probability of the next token depends on information seen many steps before. To solve this issue various modifications have been proposed, such as long-short-term-memory (LSTM) \cite{Hochreiter97} that introduce a gating mechanisms. \\

As a baseline model for comparison we trained a n-gram model. The n-gram model is a model that gives probability to each token based on how frequent did the sequence of the last $n-1$ tokens end in this token in the training set. The main limitation of n-gram models is that for small n the context used for prediction is very small, while for large $n$ most test sequences of size $n$ are never seen in the training set. We used a 2-gram model, i.e. each word is predicted according to the frequency that it appeared after the previous one. 
  \section{Results}
In order to generate our datasets we collected transliterated texts from achemenet website, based on data prepared by F. Joannès and his team in the framework of the Achemenet Program (CNRS, Nanterre). 
We then designed a tokenization method for Akkadian transliterations, as detailed in Sec. \ref{sec:data}. We trained a LSTM recurrent network and a n-gram baseline model on this dataset, see supporting information for model and training details.\newline

Results for both models are in table \ref{tab:loss}. Loss refers to mean negative log-likelihood and  perplexity is two to the power of the entropy (both cases lower is better).

\begin{table}[h]
\begin{tabular}{|c|c|c|c|c|}
\hline
       & Train loss & Train perplexity & Test loss & Test perplexity \\ \hline
2-Gram & 3.68          & 12.84               & 4.51         & 22.87              \\ \hline
LSTM   & 1.45         & 4.28              & 1.61         & 5.02             \\ \hline
\end{tabular}\caption{Loss and perplexity on Achemenet dataset}\label{tab:loss}
\end{table}

As expected, the RNN greatly outperforms the n-gram baseline and despite the limitations of the dataset it does not suffer from severe overfitting.

\subsection{Completing random missing words}
In order to evaluate our models' ability to complete missing words, we took random sentences from the test corpus, removed the middle word and tried to predict it using the rest of the sentence. Our model returns a ranking of probable words and we report the mean reciprocal rank (MRR). The MRR is the average over the dataset of one over the predicted rank of the correct word. It is a very common and useful measure for information retrieval as it is highly biased towards the top ranks, which is what the user is mostly interested in. We also evaluate the ``hit@k'' with measure what is the percentage of sentences in which the correct completion is in the top k suggestions. We took all test sentences without breaks of length 10 or longer, 166 sentences in total, for evaluation.\\

 We compared two variations of our model, one that finds the optimal completion based only on the words up until the missing word, denoted ``LSTM (start)'', and one that takes the full sentence into account labeled ``LSTM (full)''. As the ``LSTM (full)'' model needs to run separately for each candidate missing word, we first picked the top 100 candidates using ``LSTM (start)''. We then generated the 100 sentences, one for each possible completion and re-rank them based on the full sentence log-likelihood. If the right completion is not in the top 100, we take the reciprocal rank to be zero.\\

\begin{table}[h]
\centering

\begin{tabular}{|l|l|l|l|l|}
\hline
5th index & MRR & Hit@1 & Hit@5 & Hit@10 \\ \hline \hline
2-Gram (start)       & 0.51   & 32.1\%     & 75.5\%     & 82.1\%      \\ \hline
LSTM (start) & 0.756   & 66.5\%     & 87.5\%     & 91.9\%      \\ \hline
 \hline
2-Gram (full)       & 0.64  & 52.0\%     & 78.2\%     & 83.6\%      \\ \hline

LSTM (full)  & 0.89   & 85.0\%     & 94.0\%     & 95.5\%  \\    \hline
\end{tabular}
\caption{Completing missing fifth word in sentences.}
\label{tab:random}
\end{table}

 For comparison we used two simple 2-gram baselines. One that takes into account only the previous word, denoted "2-Gram (start)", and one that takes into account both previous and next word denoted "2-Gram (full)". While this is a relatively weak model, we found it to work surprisingly well yet still significantly inferior to the LSTM model.\\

It is clear from the results in table \ref{tab:random} that our algorithm can be of great help in completing missing words, with almost 85\% chance of completing the word correctly and 94\% chance of having the correct word in the top 10 suggestions.  
\subsection{Designed completion test}
We designed another experiment in order to evaluate our completion algorithm and understand its strengths and weaknesses. We generated a set of 52 multiple choice questions, where the model is presented with a sentence with one word missing as well as four possible completions and the goal is to select the correct one. The three wrong answers were designed to be wrong semantically, wrong syntactically and both. This way we can see the types of mistakes the algorithms makes. The assumption is that the learning algorithm would be more likely than a human to make semantic mistakes, but should be better than a non-expert in grammar. If that would be the case, the effectiveness of our approach as a way to assist humans would rise, as the strengths of human and machine complement each other.\\

We used our model to rank four possible restorations for each of the missing words in the 52 random sentences, selecting the one with highest likelihood we received 88.5\% accuracy (see Supporting Information for the complete list of questions and answers). Looking at the six failed completions we see that four are semantically incorrect, one is syntactically incorrect and one is both; in agreement with our hypothesis.\\
  \section{Discussion}
Further study into the different restorations of the designed completion test, taking into account the full ranking of the answers, results in some interesting patterns. The majority of restorations, 36 cases, show the algorithm best identifies either correct sentence structure or correct syntactic sequences of parts of speech based on statistical frequency of smaller syntagmatic structures. A smaller subset of cases, which probably derives from paradigmatic relationships between certain classes of words, show correct semantic identification of noun class, as well as related verbs. With regards to the latter, five possible cases identify correctly usage of verbal forms based on their context (e.g. in direct speech). Take for example question 3: NAME a\emph{šú} \emph{šá} NAME \emph{ana} NAME lú \emph{qíipi} ébabbarra \emph{u} NAME lú sanga LOCATION \textunderscore\textunderscore\textunderscore \,\,\emph{umma}. The model ranked the four possible answers as follows: \emph{iqbi}; \emph{liqbuú}; bar; bán. The example does not only show correct identification of sentence structure, but also linking two different forms of the verb \emph{qabû} "to speak". It does not necessarily reflect understanding of verbal root form, rather statistical frequency of \emph{iqbi} in this context and identification of its similarity to \emph{liqbuú}. This statistical inference emerges more clearly in one of the mistakes made by the model in question 32, where it does not differentiate properly the grammatical person of the verb \emph{nadānu} "to give, pay" (\emph{taaddinu} vs. \emph{inamdin}).\\

Some level of the models' semantic knowledge becomes apparent with regards to noun class. 16 questions show possible correct identification of countable nouns, names of professions, temporal designations, gender, and even contextual formulaic legal clause (so called \emph{elat}-clause). Six cases show correct identification of prepositions, particle use, or pronouns. The choice in question 7 between the family of related prepositions \emph{ina} and \emph{ana}, makes it clear that these choices are again based on frequency in specific contexts. Moreover, statistical grasp of parts of speech seems to be a decisive factor in at least six cases of restoration. But it can either interfere with contextual identification of the correct restoration--e.g. by preferring \emph{ina} igi over \emph{ina} $\text{šu}^{\text{II}}$ before NAME (question 35)--or achieve surprisingly good results--e.g. kurkur over LOCATION after lugal (question 37).\\

The model does not seem to identify alternate logographic and phonetic writings of the same words: e.g. Sum. da = Akk. \emph{itti} or Sum. im.dub = Akk. \emph{tuppi}. It obviously lacks enough examples of interchangeability between cases in the studied corpus. Further confusion can happen when the model identifies similarity between the answer and another word close by in the sentence, either noun or verb. Especially problematic are cases when there are very few similar sentences to train on, so the algorithm makes an "educated" guess.\\

  \section{Conclusion}
In conclusion, our model--as far as can be judged by this experiment--is, as expected, good in teasing out sentence structures. However, it was also surprisingly better than assumed in semantic identifications due to context based statistical inference (rather than finding underlying grammatical rules and morphology). In order to greatly improve false identifications based on statistical frequency of contextual semantic relationships, much more training material will be needed. Nevertheless, We have demonstrated that even without access to large amounts of data we can successfully train LSTM models and use them to complete missing words. In our completion test we show good results that while not sufficient for automatic completion, prove that this can be an invaluable tool in helping scholars with text restoration.\\

The significance of our results with the late Babylonian corpus is rooted in the fact that most entry level scholars or other interested historians and social scientists, who focus on the large first millennium BCE Babylonian archives, do not have the very specific knowledge and expertise to understand deep underlying political, social or historical structures without reading through hundreds of texts. By incorporating our model in an appropriate tool (made available on-line in the near future through the Babylonian Engine project), it will be of immense help to scholars in the historical sciences, allowing them to overcome the high entry barrier needed to restore fragmented Akkadian texts; first structured archival documents, but as the data set grows one can train the model on more genres, such as scientific or literary texts. Access to the primary sources in their original state as well as the ability to restore broken passages are a necessity for understanding Akkadian corpora on a macroscale.

  \section{Related work}

Our method is innovative in its implementation on ancient cuneiform texts. However, to better understand the significance of this study, it should be placed in the broader context of the necessary data pipeline for reading such ancient texts. One can classify two types of relevant problems in text restoration that are currently being dealt with state-of-the-art machine assisted solutions:
\renewcommand{\labelenumi}{\Roman{enumi}}
\begin{enumerate}
\item Problems of visualization which relate to the preservation, reconstruction and accessibility of documentary sources using some form of scanning, photography or both, in 2D+ or 3D technology \cite{Fisseleretal14,Pauzi17}. Nowadays, the most cost effective methods combine Photogrammetry, which creates a 3D (or 2D+) model of the object \cite{LewisChng12}, and Polynomial Texture Mapping (PTM) using Reflection Transformation Imaging (RTI) technology. The latter provides different lighting sources and texture to the scanned object \cite{Earletal11}. Some systems employ multispectral imaging that can reveal features hidden from the human eye \cite{Soberetal14}. Several major projects developed effective methods for 3D or 2D+ scans and photography of cuneiform tablets. \textit{(a)} The pioneering, but now defunct iCaly project from the University of Johns Hopkins \cite{Cohenetal04}; \textit{(b)} the PTM and RTI dome shaped systems developed by Southampton and Oxford\cite{Earletal11}, on the one hand, and by KU Leuven\cite{HameeuwWillems11}, on the other; \textit{(c)} a joint Dortmund-Würzburg team that scans cuneiform fragments in 3D and focuses on digitizing philological work and reconstruction of fragmentary tablets\cite{Fisseleretal14}; \textit{(d)} and the initiative GigaMesh, led by Hubert Mara from Heidelberg \cite{Maraetal10}. The Heidelberg group have recently developed various methods for Automatic Machine identification of cuneiform signs using ML models and Vector Geometry \cite{Bogaczetal15,BogaczMara17}. 
Advances in cost-effective and fast 3D scanning technology are crucial to further the work described here. For instance, they allow exact measurements of inscribed objects, that can lead to the joining of broken tablet fragments. These can otherwise only be identified as matching by a handful of world experts in cuneiform. The Virtual Cuneiform Tablet Reconstruction Project (\href {http://virtualcuneiform.org/}{VCTR}) joined 3D scanned cuneiform tablet fragments automatically using a novel matching algorithm with measure of fit metrics which dramatically reduce false positive match reports \cite{Collinsetal14}. The matching algorithm works by iteratively finding the optimal relative orientation of the two fragments under consideration in three-dimensional space. The team succeeded in joining with this method Neo-Babylonian archival texts as well as a manuscript of the Babylonian flood myth Atrahasis \cite{Collinsetal17}.
\item Linguistic and content-related problems, which include automated or partly automated transcription and translation of ancient languages. This is an area with potential for Big data mining using models of Natural Language Processing (NLP), ML or AI. It is also the most complicated aspect, given the lexical and semantic complexity of the cuneiform script and Akkadian language. A multi-national project led by a group from Toronto, Frankfurt and UCLA has initiated the Machine Translation and Automated Analysis of Cuneiform Languages project (\href {https://cdli-gh.github.io/mtaac/}{MTAAC}). Its main goal is to find methodologies for the automated analysis and machine translation of transliterated cuneiform documents, specifically written in the less syntactically complex Sumerian language \cite{PagePerronetal17}. They aim to have the resulting translated lemmas automatically tagged according to context, creating a semantic and lexical database, based on neural machine translation models.
A recent endeavor of a joint Ariel-Tel Aviv research group, managed high success rates using NLP algorithms like HMM, MEMM and BiLSTEM, for word segmentation and automatic transliteration of Akkadian texts in Unicode cuneiform \cite{Gordinforth}.
\end{enumerate}
  \section{Materials and Methods}\label{sec:data}

\subsection*{Transcription of Akkadian Cuneiform script and its Neo-Babylonian dialect}
\begin{figure}
  \includegraphics[width=0.5\textwidth]{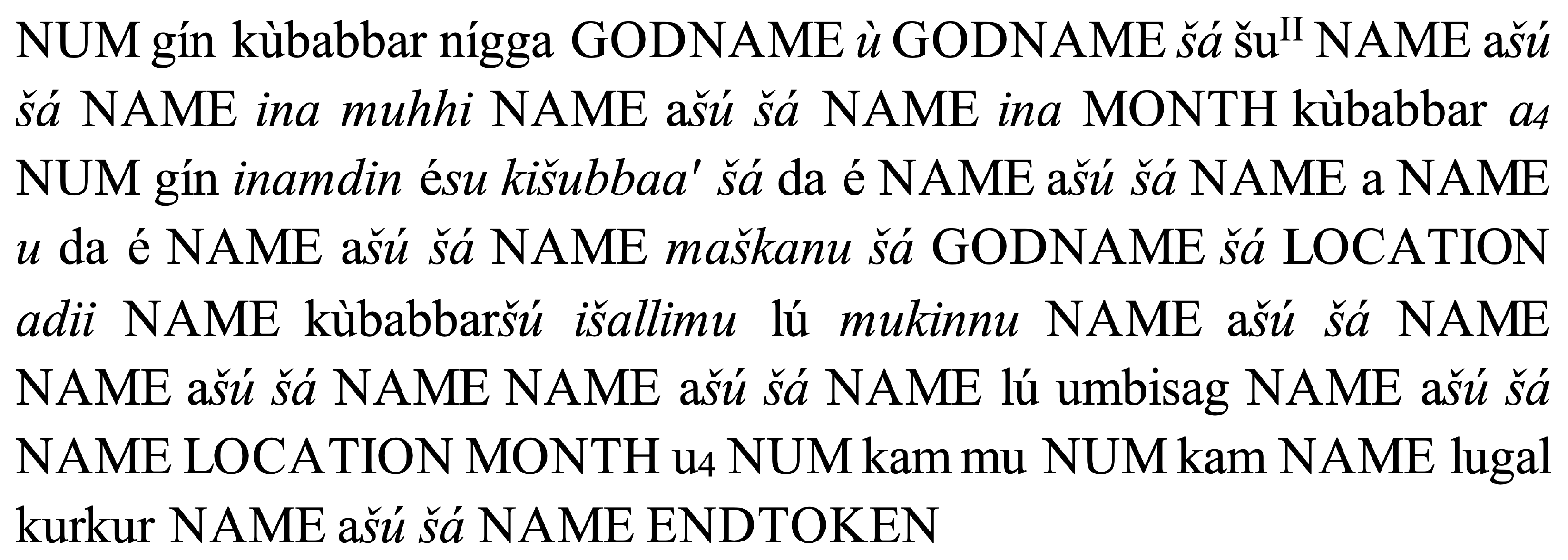}
  \caption{Mechanical bound transcription of Babylonian text YOS 7 11.}\label{Fig:example1}
\end{figure}

  Akkadian was written in the Cuneiform script. Alongside Egyptian Hieroglyphs, Cuneiform is the earliest attested form of writing, which was probably invented in southern Mesopotamia at the end of the fourth Millennium BCE and initially used to record daily accounting procedures in the Sumerian cities on a clay medium. A good analogy to this earliest phase is the modern "spreadsheets" see \cite{Veldhuis12}. The script was then adopted by the Akkadian speaking Babylonians and Assyrians to write their own language using a mixture of syllabic signs, logograms (which incorporated Sumferian values for ideograms) and determinatives\cite{HuehnergardWoods04}. In all, Akkadian is one of the most enduring and widely attested languages of the ancient Near East for around 2,500 years. Its geographic horizon spans from Iran to Greece and from Anatolia to Egypt.
  Neo-Babylonian is the longest consecutive language phase of Akkadian, covering the first millennium BCE, ending sometime after the first century CE. The genres and writing conventions of this phase are characterized by their departure from standardized orthography practiced throughout the second millennium BCE. Many spellings are inconsistent with actual phonemic renderings of words and can vary to a considerable extent,\footnote{Take for example the form of a very common word in the Nippur Achaemenid period Murašu archive \textit{ha\d{t}ru}. As shown by Stolper\cite{EE}, the different spellings of this term leave the quality of the middle, dental consonant uncertain: (lú) \textit{ha-ad/t/ \d{t}-ru/ri}, its variants range from (lú) \textit{ha-d/ \d{t}a-ri}, (lú) \textit{ha-dar/tár/ \d{t}ár}, and (lú) \textit{ha-d/ \d{t}a-ad/t/ \d{t}-ri}} especially on account of the intensive language contact and interference between Akkadian and Aramaic\cite{Streck03,Streck10a}. There are some rules that govern the normalization of Neo-Babylonian--i.e. bound transcription which correctly represents noun and verbal morphology--but in general it is avoided in most recent publications unless for linguistic or pedagogic purposes\cite{Jursa05}. For this reason we have also chosen to avoid the pitfall of training the algorithm in any kind of normalization practices for the time being. In our training corpus we remained on the level of (unbiased) transliteration, but we removed all connecting features between phonograms and sumerograms, resulting in a mechanical (un-normalized) bound transcription: Akkadian phonetic spellings and logographic writings are taken at face value, by simply removing connecting hyphens between syllables and between logograms. A necessary contrast is drawn between phonetic and logographic writings based on their typical representation in italic typeface vs regular typeface, respectively (see also below on Tokenization). However, phonetic compliments, normally italicized, are currently identified as part of regular typeface logograms when attached. Superscripted determinatives are used to identify proper names, such as personal names, theophoric names, locations and month names. Changes in sentence structure were not taken into consideration, since they only occur at the relatively late corpora of the Parthian Period (third century BCE onwards)\cite{Hackl18}.
  The resulting Akkadian texts used to train the algorithm look like the example in Fig. \ref{Fig:example1}, which show the same text as in Fig. \ref{Fig:example} but in our mechanical bound transcription.
  
\subsection*{Neo-Babylonian archives under the Persian Empire, their historical significance and text restorations}
  Babylonian archives from the end of the sixth to the fourth centuries BCE are one of the main sources for reconstructing the official and daily heritage of the Persian Empire and its subject peoples in Mesopotamia. Structuring of Neo-Babylonian archives is based mostly on an artificial division between private and institutional ownership\cite{Jursa10}. Criteria employed to this end are more frequently reliant on common principal actors with connected activities (i.e. prosopography), document type and content or shared background in an institution (like temple or palace), and less on physical proximity between documents in a given find context (archaeological and/or museum based studies in acquisition history for illicitly excavated texts). Among the largest representative text groups with a private background are the business archives of the Egibi and Nūr-Sîn families from Babylon and Murašu 'firm' from Nippur, as well as the closely contemporary archive of Persian governor Bēlšunu from the palace complex of Babylon, known as the Kasr archive (designated Kasr N6; \cite{Pedersen05}).\footnote{Kasr has in fact a mixed private and institutional background, see \cite{Waerzeggers18} for an overview of cuneiform archives from Achaemenid period Babylonia and their time span. A more detailed discussion of each text group is found in \cite{Jursa05}} The Murašu texts especially and another archive cluster from several rural centres known as the Yahudu 'archive', provide significant information on foreign minority communities in the Achaemenid Empire during a period of close to 200 years, including the fate of the Judean community in Babylonian exile\cite{PearceWunsch14}. However, the largest textual groups from this period by far are the two large multi-file archives with an institutional background in city temples: the Eanna archive from Uruk and Ebabbar archive from Sippar. These makeup the bulk of the Achemenet data set, alongside the Egibi/Nūr-Sîn archive and Murašu material. All together, the Achemenet Neo-Babylonian data set has representative archival groups for the Achaemenid period from almost every large city in Babylonia:\footnote{Archives from Borsippa and Ur are not yet attested. Designations of archives are in parenthesis following each city name. Archives already mentioned above (like Murašu from Nippur) are not included in this list} Babylon (Ea-eppēš-ilī, Gahal, Nappāhu); Kiš (Eppēš-ilī); Sippar (Bēl-rēmanni, Ea-eppēš-ilī A, Iššar-tarībi, Marduk-rēmanni, Rē'i-sisê); Uruk (Atû).
  
\begin{figure*}
  \includegraphics[width=17.8cm]{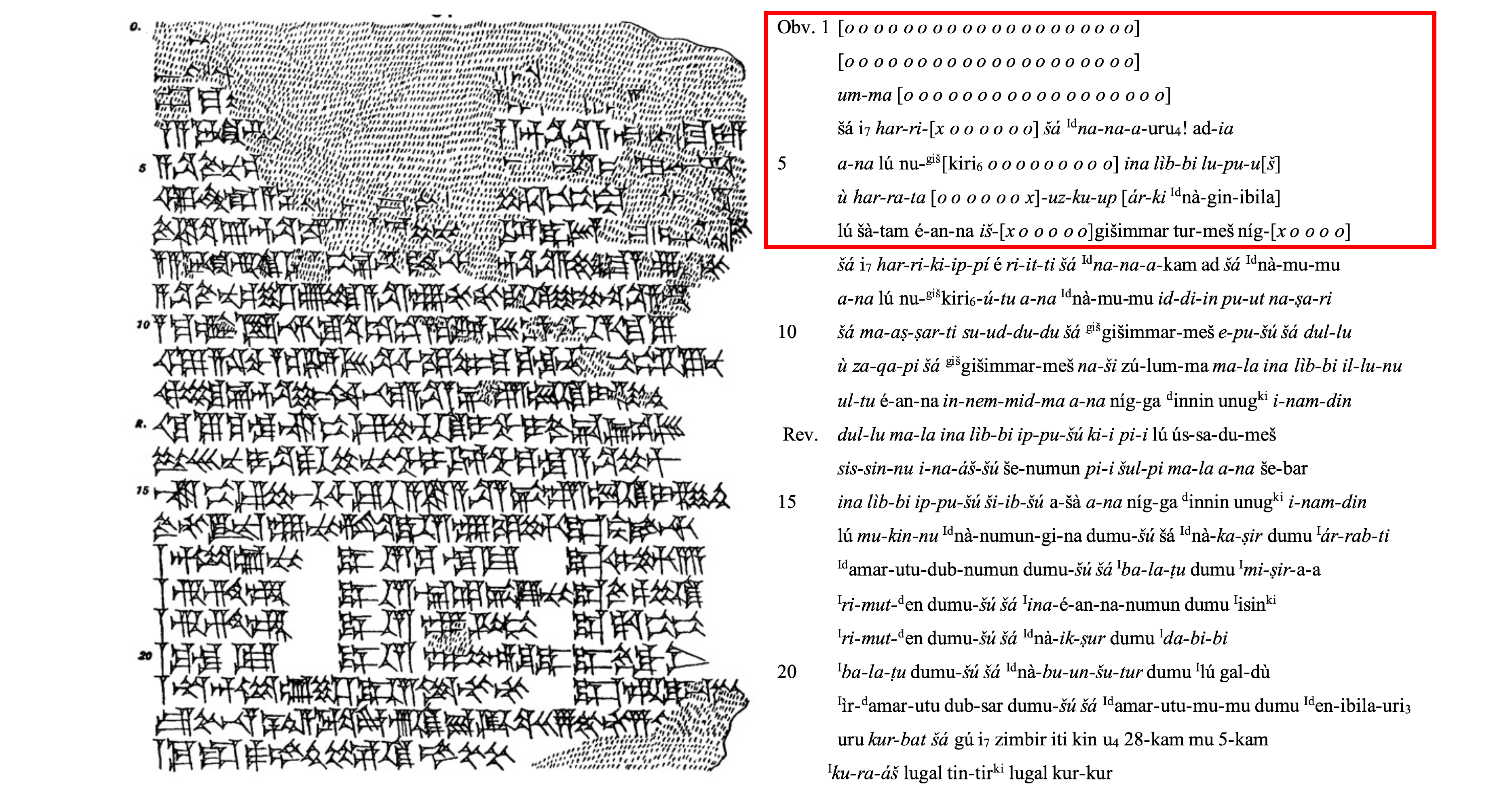}
  \caption{Line art and transliteration of Achaemenid period Babylonian text YOS 7 51 from the Eanna archive in Uruk. Fragmentary upper half of obverse marked by a red square.}\label{Fig:example2}
\end{figure*}

  The need for text restorations varies from archive to archive depending either on their method of excavation and preservation in recent times, or archival selection processes in antiquity (e.g. discarded or "dead" archives). The best kept tablets found their way into Museum collections in Europe and the US already following their initial age of discovery during the late 19th and early 20th century. Many came from illicit or clandestine excavations and went through an active selection process, by which collections preferred the most complete tablets. On the other hand, tablets from official excavations in Babylon and and Uruk, for example, have a higher percentage of fragmentary texts. Some large archives like Murašu or Kasr (that was already vitrified from an ancient fire), became damaged because of poor handling following excavation or due to the effects of war\footnote{The Murašu texts were damaged during their transport out of Nippur, see \cite{EE}, and Kasr texts partially survived a grim sequence of events triggered by the first World War. Nevertheless, many of them already suffered ancient fire damage in during or after the Achaemenid period \cite{Stolper07}}. A large number of Eanna tablets from before the reign of Darius I were deliberately discarded or smashed already in antiquity after becoming inactive for the temple administration\cite{YOS21,Kessler18}. Such an Eanna text with a fragmentary upper half of the obverse, dating to the reign of Cyrus can be seen in Fig. \ref{Fig:example2}, followed by its possible restoration that is based on known parallels and scholarly study (Fig. S1).

\subsection*{Data collection}
We collected 2,247 Achaemenid period Babylonian archival texts. As the Achemenet website does not have an API we built a scraping code in Python 2.7 to scrap the texts, preprocessing and tokenize them. The code uses the "Beautiful Soup"  library to remove all the the unnecessary HTML tags and take only the transliterated text itself from the site. Superscript and italic tags have semantic meaning and were therefore preserved by our processing. We replace words with low appearance (below three total appearances in the train data) with an UNKNOWN token, as there is not enough data to properly predict and use these words. 
  The number of different words in the vocabulary that were collected after this is 1,549 and the number of words in total is 220,926. The number of words that appear only once is 3,175 and twice is 932. For comparison the Penn treebank dataset, a standard and relatively small English text data set, comprising of texts from Wall Street Journal, has 10,000 unique words and the number of total words is 1,036,580. While over-fitting is something to be aware of given the scale of the data, the unique nature of these texts comprising of well structured bureaucratic information makes them well suited for machine learning modelling.    

\subsection*{Tokenization}
  Tokenization is an automatic process in which the text is split into words, and each one is replaced by a numeric token. This is an important process that requires language specific knowledge, or a lot of semantic meaning might be lost. A classical example in English is tokenizing a word like ``aren't". If we do not break it into two tokens then it is considered a word on its own and losses the connection to ``are" and ``not". While it might be possible for the learning algorithm to learn the connection from the data, bad tokenization can complicate matters considerably by creating a large amount of unnecessary words in our dictionary.\newline
  
  We created a new tokenizer, specifically built for Akkadian. Masculine names, god names and female names, identified by determinatives before proper names in super script 'I' or 'Id', 'd' and 'f' respectively, were replaced by a 'NAME', 'GODNAME' and 'FEMALENAME' token. Locations, identified by determinative before proper names in super script 'uru', or after proper names in super script 'ki', were replaced by 'LOCATION' token. Month names, with determinative super script 'iti' before the noun, were replaced by MONTH and simple numbers were replaced with the token 'NUM'. In order to simplify the tokenization of broken parts, each broken or incomplete part was replaced with the token '\textless BRK\textgreater', and words that appeared only two times or less as '\textless UNK\textgreater', since we we do not have enough information to learn their meaning.\newline 
  
  Another important aspect of Akaddian is that some cuneiform symbols can be interpreted in two ways: As a syllable or as a logogram, i.e. representing a whole word. During transliteration the specific meaning is marked by using italic for syllables. During tokenization we use the same token for both representations, but we keep the HTML start italic <i> and stop italic <\textbackslash i> symbols so the use of the word as a syllable or logogram can be inferred by the context. While using different tokens for both uses has some advantages, we found that doing so adds a large amount of noise to the preprocessing step and decided to use this method instead. For example, one sentence fragment after preprocessing is "NUM mana kùbabbar <i> šá </i> NAME a <i> šú šá </i> NAME". After this preprocessing every unique word is replaced by a numeric token.

\showmatmethods{} 

\acknow{The research reported here received funding from the Ministry of Science and Technology Grant 89540 and the Israel Science Foundation Grant 457/19.}

\showacknow{} 
https://www.overleaf.com/project/5bbd570d71590a2759027677
\bibliography{pnas_bib}

\end{document}